\documentclass[lettersize,journal]{IEEEtran}
\usepackage{algorithmic}
\usepackage{multirow}
\usepackage[numbers,compress]{natbib}
\usepackage{xspace}
\usepackage{amsmath,amsfonts}
\usepackage{algorithmic}
\usepackage{algorithm}
\usepackage{array}
\usepackage[caption=false,font=normalsize,labelfont=sf,textfont=sf]{subfig}
\usepackage{textcomp}
\usepackage{stfloats}
\usepackage{url}
\usepackage{balance}
\usepackage{verbatim}
\usepackage{amsthm}
\newtheorem{definition}{Definition}
\usepackage{graphicx}

\makeatletter
\def\hlinewd#1{%
	\noalign{\ifnum0=`}\fi\hrule \@height #1 \futurelet
	\reserved@a\@xhline}
\makeatother
\newcommand{\sys}{\textsc{\textsf{HeteroSample}}\xspace}

\begin{document}

\title{\sys: Meta-path Guided Sampling for Heterogeneous Graph Representation Learning}

\author{Ao Liu$^1$, Jing Chen$^1$, Ruiying Du$^1$, Cong Wu$^3$, Yebo Feng$^3$, Teng Li$^2$, Jianfeng Ma$^2$\\
$^1$\emph{Wuhan University}, $^2$\emph{Xidian University, China}, $^3$\emph{Nanyang Technological University, Singapore}\\
\scriptsize \texttt{\{liuao6,chenjing,duraying\}@whu.edu.cn, 
	\{cong.wu,yebo.feng\}@ntu.edu.sg,
	litengxidian@gmail.com, jfmajfma@mail.xidian.edu.cn}}

\maketitle

\begin{abstract}
	The rapid expansion of Internet of Things (IoT) has resulted in vast, heterogeneous graphs that capture complex interactions among devices, sensors, and systems. Efficient analysis of these graphs is critical for deriving insights in IoT scenarios such as smart cities, industrial IoT, and intelligent transportation systems. However, the scale and diversity of IoT-generated data present significant challenges, and existing methods often struggle with preserving the structural integrity and semantic richness of these complex graphs. Many current approaches fail to maintain the balance between computational efficiency and the quality of the insights generated, leading to potential loss of critical information necessary for accurate decision-making in IoT applications.
	We introduce \sys, a novel sampling method designed to address these challenges by preserving the structural integrity, node and edge type distributions, and semantic patterns of IoT-related graphs. \sys works by incorporating the novel  top-leader selection, balanced neighborhood expansion, and meta-path guided sampling strategies. The key idea is to leverage the inherent heterogeneous structure and semantic relationships encoded by meta-paths to guide the sampling process. This approach ensures that the resulting subgraphs are representative of the original data while significantly reducing computational overhead. Extensive experiments demonstrate that \sys outperforms state-of-the-art methods, achieving up to 15\% higher F1 scores in tasks such as link prediction and node classification, while reducing runtime by 20\%.These advantages make \sys a transformative tool for scalable and accurate IoT applications, enabling more effective and efficient analysis of complex IoT systems, ultimately driving advancements in smart cities, industrial IoT, and beyond.
\end{abstract}


\section{Introduction}

The Internet of Things (IoT) has revolutionized the way we model and interact with complex real-world systems~\cite{lin2024efficient,lin2024fedsn,lin2023pushing}, 
generating vast, heterogeneous graphs where diverse types of nodes and edges represent the intricate relationships between interconnected devices, sensors, and systems~\cite{wang2022survey,fu2020magnn,yang2020heterogeneous,liu2024dynashard}. These heterogeneous graphs have become essential tools for IoT applications such as smart cities, industrial automation, and intelligent transportation~\cite{lin2024adaptsfl,fang2024ic3m,lin2024split,lin2022channel}, where understanding the complex interactions between entities is critical~\cite{liu2019log2vec,wang2021polymorphic,wu2023heterogeneous,zhao2022cyber}. Effective representation learning of these graphs is vital for a variety of downstream tasks, including node classification, link prediction, and anomaly detection, which are crucial in maintaining the functionality and security of IoT networks~\cite{zhang2023graph,wang2023multi,fan2021heterogeneous,luo2021detecting}. However, the scale and complexity of IoT-generated heterogeneous graphs present significant challenges, particularly in terms of computational efficiency and scalability. Existing graph sampling techniques aim to address these challenges by extracting representative subgraphs to reduce size~\cite{wu2020comprehensive,huang2018adaptive,zou2019layer,zhao2020preserving}, but they often struggle to preserve the rich semantic and structural information necessary for accurate IoT data analysis.

\textbf{Existing work.} Existing approaches to heterogeneous graph sampling in IoT scenarios can be broadly classified into two main categories: random walk-based methods and node-wise sampling methods. Random walk-based methods, such as meta-path based random walks~\cite{wang2020dynamic,jiang2017semi} and heterogeneous graph attention networks (HGAT)\cite{yang2021hgat}, are designed to traverse the graph along various meta-paths, capturing the diverse semantic and structural information embedded within the graph. These methods have been effective in several contexts but often encounter significant computational complexity, making them less suitable for the large-scale and dynamic nature of IoT-generated data. On the other hand, node-wise sampling methods, like HGSampling\cite{ma2023single} and HetGNN~\cite{ansarizadeh2023deterministic}, focus on selecting nodes based on their importance or centrality measures, attempting to preserve the key structural features of the graph. HGSampling uses a budget-based strategy to select informative nodes, while HetGNN employs a heterogeneous graph neural network to learn detailed node representations.

\textbf{Research gap.} Despite the advancements these methods offer, significant challenges remain, particularly in the context of IoT applications. Random walk-based methods often struggle to maintain the global properties of the graph, which are essential for accurately modeling the interconnected systems typical in IoT environments. Similarly, node-wise sampling approaches, while effective in preserving certain structural features, often fall short in capturing the complex interactions between different node and edge types. This limitation becomes particularly pronounced in IoT scenarios, where the accurate representation of intricate relationships is crucial for tasks such as anomaly detection, predictive maintenance, and network optimization. These gaps highlight the need for more robust and efficient sampling methods that can better preserve the structural integrity and semantic richness of heterogeneous graphs in IoT contexts.

\textbf{\sys.} To address the limitations of existing methods, we propose \sys, a novel deterministic sampling method specifically designed for heterogeneous graphs. \sys is crafted to preserve both the structural and semantic properties of these graphs while maintaining computational efficiency and scalability. The core idea of \sys is to utilize a deterministic approach that leverages both local and global information to select representative nodes within the graph. Unlike traditional methods that depend on random walks or node-wise sampling based solely on centrality measures, \sys employs a three-step process. First, it identifies the most influential nodes for each node type by analyzing their centrality scores, ensuring a comprehensive representation of the graph's heterogeneity. Second, it expands the neighborhoods of these selected top-leaders in a balanced manner, ensuring a proportional representation of different node types within the local context. Lastly, it incorporates meta-path guided expansion to capture and preserve the semantic relationships among nodes, leading to a more informed and effective sampling process. Through these steps, \sys aims to maintain the essential properties of the heterogeneous graph while significantly reducing its size, striking an optimal balance between effectiveness and efficiency.

\sys provides several advantages over existing methods. Firstly, it effectively preserves the structural properties and inherent heterogeneity of the original graph by considering both node importance and a balanced representation of various node types. Secondly, \sys captures the semantic relationships among nodes by leveraging meta-paths during the sampling process, allowing for a more comprehensive and accurate representation of the graph. Lastly, \sys is both computationally efficient and scalable, focusing on selecting representative nodes and their immediate neighborhoods without relying on costly global computations. These advantages make \sys a powerful tool for analyzing large-scale heterogeneous graphs, particularly in data-intensive fields such as IoT.

Integrating top-leader selection with meta-path-based expansion presents several technical challenges, particularly in balancing local and global graph features. Top-leader selection focuses on identifying influential nodes based on local connectivity, while meta-path-based expansion aims to capture the broader semantic relationships among diverse node types. During implementation, we faced challenges such as ensuring that the selected top-leaders effectively represented local neighborhoods while also preserving the overall semantic integrity of the graph. To address these issues, we developed a hybrid approach that leverages the strengths of both methods, allowing for a comprehensive sampling strategy that maintains both structural characteristics and rich semantic patterns. By tackling these technical challenges, our method not only enhances the representativeness and quality of the sampled subgraphs but also demonstrates significant robustness in the field of heterogeneous graph analysis.

Our main contributions are summarized as follows:
\begin{itemize}
	\item We propose \sys, a novel deterministic sampling method for heterogeneous graphs that effectively preserves the structural and semantic properties of the original graph.
	\item We design a top-leader selection strategy identifying the most influential nodes of each node type,
	      a balanced neighborhood expansion approach maintaining a balanced representation of different node types,
	      and a meta-path guided expansion to capture the semantic relationships among nodes.

	\item We conduct extensive evaluations on three real-world heterogeneous graph datasets to evaluate the effectiveness of \sys in preserving the key properties of the original graph and its performance in downstream tasks of link prediction. We also compare \sys with state-of-the-art baselines and show its superiority in terms of both effectiveness and efficiency.
\end{itemize}

\section{Problem Formulation}

In this section, we define the problem of sampling heterogeneous graphs and introduce the notations and definitions.

\subsection{Definitions}

\begin{figure}
	\centering
	\includegraphics[width = 0.95\linewidth]{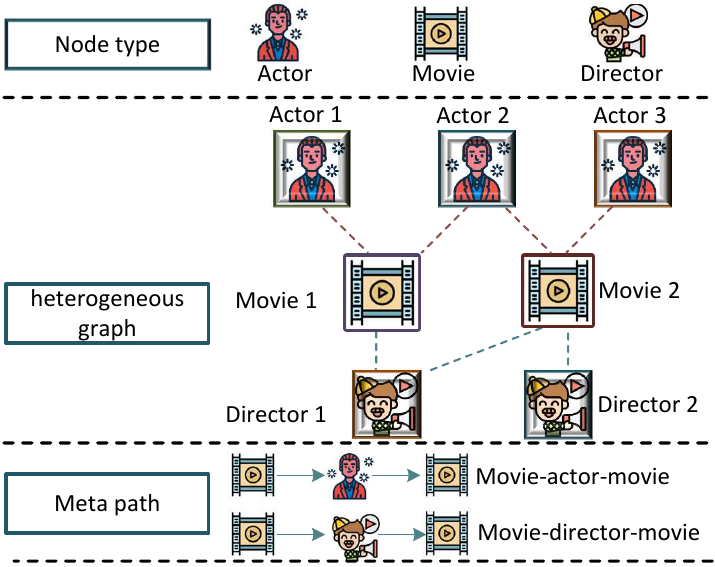}
	\caption{Illustration of heterogeneous graph}
	\label{fig:heter_graph}
	\vspace{-5mm}
\end{figure}

\begin{definition}
	\textbf{Heterogeneous Graph}. As illustrated in Figure~\ref{fig:heter_graph}, a heterogeneous graph consists of various types of nodes and edges, defined as a graph \( G = (V, E, \phi, \psi) \). Here, \( V \) represents the set of nodes, while \( E \) denotes the set of edges. The mapping functions \( \phi: V \rightarrow \mathcal{A} \) and \( \psi: E \rightarrow \mathcal{R} \) categorize nodes and edges into their respective types. Specifically, \( \mathcal{A} \) and \( \mathcal{R} \) represent the sets of node types and edge types, respectively. We denote the set of node types as \( \mathcal{A} = \{A_1, A_2, \ldots, A_m\} \) and the set of edge types as \( \mathcal{R} = \{R_1, R_2, \ldots, R_n\} \). The interactions among these diverse types are critical for our sampling process, as they allow us to effectively capture the complex relationships encoded in the graph, thereby enhancing the representativeness of the sampled subgraphs.
\end{definition}

\begin{definition}
	\textbf{Meta-path}. A meta-path $P$ is defined as a path in the form of $A_1 \xrightarrow{R_1} A_2 \xrightarrow{R_2} ... \xrightarrow{R_l} A_{l+1}$, which describes a composite relation between node types $A_1$ and $A_{l+1}$, where $A_i \in \mathcal{A}$ and $R_i \in \mathcal{R}$ are node types and edge types, respectively.
\end{definition}

\begin{definition}\textbf{Edge Type Importance}. The edge type importance matrix \( \boldsymbol{W} = (w_{ij})_{m \times m} \) assigns an importance weight to each edge type, where \( w_{ij} \) represents the importance of the edge type connecting node types \( A_i \) and \( A_j \). In practice, these weights can be determined using domain knowledge or through frequency analysis of the interactions in the dataset. For instance, if certain edge types consistently connect highly influential node types, they may be assigned higher weights based on their observed frequency of occurrence and significance in previous analyses. The importance weights satisfy the condition \( \sum_{i=1}^{m} \sum_{j=1}^{m} w_{ij} = 1 \), ensuring that the weights are normalized. It is important to note that the importance of edge types may vary across different datasets, which could affect the generalizability of the method. Thus, adapting the weighting strategy based on specific dataset characteristics is essential for optimizing performance.
\end{definition}

\begin{definition}
	\textbf{Edge Type Importance}. The edge type importance matrix $\boldsymbol{W} = (w_{ij})_{m \times m}$ assigns an importance weight to each edge type, where $w_{ij}$ represents the importance of the edge type connecting node types $A_i$ and $A_j$. The importance weights satisfy $\sum_{i=1}^{m} \sum_{j=1}^{m} w_{ij} = 1$.
\end{definition}

\begin{definition}
	\textbf{Meta-path Importance}. The meta-path importance vector $\boldsymbol{\beta} = (\beta_1, \beta_2, ..., \beta_q)$ assigns an importance weight to each meta-path, where $\beta_i$ represents the importance of meta-path $P_i \in \mathcal{P}$. The importance weights satisfy $\sum_{i=1}^{q} \beta_i = 1$.
\end{definition}

\begin{figure}[!t]
	\centering
	\includegraphics[width = \linewidth]{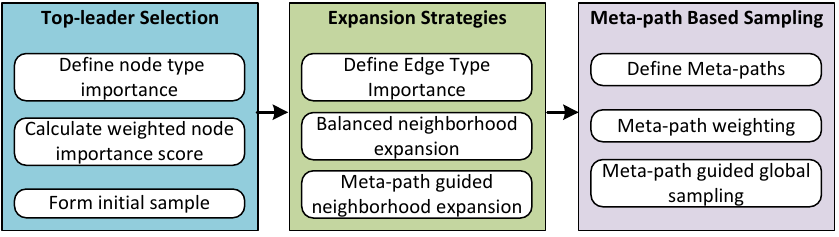}
	\caption{Workflow of \sys}
	\label{fig:overview}
	\vspace{-6mm}
\end{figure}
\subsection{Problem Statement}

Given a heterogeneous graph \( G = (V, E, \phi, \psi) \), the goal of heterogeneous graph sampling is to select a subgraph \( S = (V_S, E_S) \), where \( V_S \subseteq V \) and \( E_S \subseteq E \), that preserves key properties of \( G \), such as node type distribution, edge type distribution, meta-path-based patterns, and clustering structure, while minimizing the size of \( S \). To achieve this, \sys leverages node type importance, edge type importance, and meta-path importance to guide the sampling process, ensuring that the selected subgraph effectively captures the structural and semantic characteristics of the original graph. The proposed algorithm addresses the challenges of sampling heterogeneous graphs by maintaining their inherent properties and providing a compact representation suitable for downstream tasks.

\section{Design of \sys}
In this section, we present the overview of \sys, and detail the design of \sys.
\subsection{Overview}
\sys consists of three main steps: top-leader selection, expansion strategies, and meta-path-based sampling. In the top-leader selection step, it identifies influential nodes of each type based on weighted importance scores that consider both node degree and type significance. The expansion strategies, including Balanced Neighborhood Expansion (BNE) and Meta-path Guided Neighborhood Expansion (MGNE), add peripheral nodes from each type in a balanced manner and prioritize nodes that form important meta-path patterns. The meta-path-based sampling further refines the graph by capturing semantic relationships between different node types. By integrating these strategies, \sys effectively preserves the rich semantics and structural properties of the original graph while reducing its size, ensuring balanced representation and capturing essential characteristics. This comprehensive framework makes \sys suitable for various downstream tasks, such as node classification, link prediction, and community detection in diverse graph mining applications.

\subsection{Top-leader Selection}
In heterogeneous graphs, different node types may have varying importance within the overall graph structure. Therefore, it is crucial to consider both node degree and node type importance when selecting top-leader nodes. By initializing top-leaders based on a weighted combination of these factors, we aim to capture the most influential nodes within each node type, ensuring a diverse and representative initial sample.

\textbf{Step 1. Define node type importance.} We introduce a node type importance vector \( \boldsymbol{\alpha} = (\alpha_1, \alpha_2, \ldots, \alpha_m) \), where \( \alpha_i \) represents the importance weight of node type \( A_i \). These importance weights can be determined through domain knowledge or learned from the graph structure using techniques such as node embedding or centrality measures. The node type importance vector satisfies the condition \( \sum_{i=1}^{m} \alpha_i = 1 \).

\textbf{Step 2. Calculate weighted node importance score.} For each node \( v \in V \), we calculate a weighted node importance score that combines node degree and node type importance. The weighted node importance score \( I(v) \) is defined as:
\begin{align}
	I(v) = \alpha_{\phi(v)} \cdot \deg(v)
\end{align}
where \( \deg(v) \) is the degree of node \( v \), and \( \alpha_{\phi(v)} \) is the importance weight of the node type \( \phi(v) \) to which node \( v \) belongs.

\textbf{Step 3. Select top-leaders for each node type.} For each node type \( A_i \in \mathcal{A} \), we select the top-\( k \) nodes with the highest weighted node importance scores as the top-leaders. The value of \( k \) can be determined based on several factors, including the desired sample size, the number of node types, and the overall graph density. Additionally, we may employ thresholds or heuristics to ensure that only nodes meeting specific criteria for influence are selected, which helps balance the representation across various node types.

\textbf{Step 4. Form initial sample.} The initial sample \( S \) is formed by including all selected top-leaders from each node type. Formally, the initial sample is defined as:
\begin{align}
	S = \bigcup_{i=1}^{m} L_i
\end{align}
The size of the initial sample is \( |S| = m \cdot k \), where \( m \) is the number of node types and \( k \) is the number of top-leaders selected per node type.

By initializing the top-leaders based on the weighted node importance score, we ensure that the most influential nodes within each type are included in the initial sample, effectively capturing both local connectivity through node degree and the global significance of different node types. This top-leader selection lays the groundwork for subsequent expansion strategies, which grow the sample by incorporating peripheral nodes around the selected leaders. The choice of top-leaders is critical for determining the quality and representativeness of the final sampled graph. Future evaluations will explore the impact of varying \( k \) on performance, providing insights for optimizing the top-leader selection process.

\subsection{Expansion Strategies}

After initializing the top-leader nodes for each node type, the next step is to expand the sample by adding peripheral nodes around the top-leaders. The intuition is to capture the local neighborhood structure and the semantic relationships between different node types. By incorporating node types and edge types into the expansion process, we can ensure that the sampled graph preserves the heterogeneous nature of the original graph.
We design two expansion strategies: (1) balanced neighborhood expansion (BNE) and (2) meta-path guided neighborhood expansion (MGNE). Both strategies utilize the edge type importance matrix to prioritize the selection of peripheral nodes based on the importance of the connecting edge types.

\textbf{Step 1. Define edge type  importance}.
We introduce an edge type importance matrix $\boldsymbol{W} = (w_{ij}){m \times m}$, where $w{ij}$ represents the importance weight of the edge type connecting node types $A_i$ and $A_j$. The importance weights can be determined based on domain knowledge or learned from the graph structure using techniques such as edge embedding or frequency analysis. The edge type importance matrix is symmetric, i.e., $w_{ij} = w_{ji}$, and satisfies the condition $\sum_{i=1}^{m} \sum_{j=1}^{m} w_{ij} = 1$.

\textbf{Step 2. BNE.}
For each top-leader node \( v \in S \) and each connected node type \( A_i \), we calculate the number of peripheral nodes to be added as \( n_i = \left\lfloor\frac{|N_{A_i}(v)|}{\sum_{j=1}^{m} |N_{A_j}(v)|} \cdot \delta\right\rfloor \), where \( N_{A_i}(v) \) is the set of neighboring nodes of \( v \) belonging to node type \( A_i \), and \( \delta \) is the expansion factor determining the number of peripheral nodes to be added per top-leader node. BNE is essential because it ensures that node types are proportionally represented in the expanded neighborhood, preventing biases that may arise from uneven sampling. Without BNE, certain node types could be underrepresented, leading to a loss of critical information and skewing the results of downstream tasks. We then select \( n_i \) peripheral nodes from \( N_{A_i}(v) \) based on the importance of the connecting edge type, with the probability of selecting a peripheral node \( u \in N_{A_i}(v) \) being proportional to the importance weight \( w_{\phi(v),i} \) of the edge type connecting \( v \) and \( u \). This method contrasts with purely random expansion strategies, which may neglect the structural significance of node types, potentially compromising the richness of the sampled graph. The selected peripheral nodes are then added to the sample \( S \).

For each top-leader node $v \in S$ and each connected node type $A_i$, we calculate the number of peripheral nodes to be added as $n_i = \left\lfloor\frac{|N_{A_i}(v)|}{\sum_{j=1}^{m} |N_{A_j}(v)|} \cdot \delta\right\rfloor$, where $N_{A_i}(v)$ is the set of neighboring nodes of $v$ belonging to node type $A_i$, and $\delta$ is the expansion factor determining the number of peripheral nodes to be added per top-leader node. We then select $n_i$ peripheral nodes from $N_{A_i}(v)$ based on the importance of the connecting edge type, with the probability of selecting a peripheral node $u \in N_{A_i}(v)$ being proportional to the importance weight $w_{\phi(v),i}$ of the edge type connecting $v$ and $u$. The selected peripheral nodes are added to the sample $S$.

\textbf{Step 3. MGNE}.
For each top-leader node $v \in S$, we expand the sample by identifying the set of meta-paths $\mathcal{P}v$ originating from $v$, considering a predefined maximum length $l$. We then calculate the importance score $I_P$ for each meta-path $P \in \mathcal{P}v$ as $I_P = \prod{i=1}^{l} w{\phi(v_{i}),\phi(v_{i+1})}$, where $v_i$ and $v_{i+1}$ are consecutive nodes in the meta-path $P$, and $w_{\phi(v_{i}),\phi(v_{i+1})}$ is the importance weight of the edge type connecting their corresponding node types. The top-k meta-paths with the highest importance scores from $\mathcal{P}_v$ are selected, and for each selected meta-path $P$, the peripheral nodes along the meta-path are added to the sample $S$. This approach prioritizes the selection of peripheral nodes based on their contribution to completing high-importance meta-paths in the neighborhood of the top-leader node.


\begin{algorithm}[!t]
	\scriptsize
	\caption{\textbf{\sys}}
	\label{alg:heterosample}
	\begin{algorithmic}[1]
		\REQUIRE Heterogeneous graph $G = (V, E, \phi, \psi)$, node types $\mathcal{A}$, edge types $\mathcal{R}$, node type importance vector $\boldsymbol{\alpha}$, edge type importance matrix $\boldsymbol{W}$, meta-paths $\mathcal{P}$, meta-path importance weights $\boldsymbol{\beta}$, number of top-leaders per node type $k$, expansion factor $\delta$, maximum meta-path length $l$
		\ENSURE Sampled graph $S$

		\STATE Initialize $S \gets \emptyset$

		\STATE // \textbf{Step 1: Top-leader selection}
		\FOR{each node type $A_i \in \mathcal{A}$}
		\STATE $L_i \gets \emptyset$
		\FOR{each node $v \in V$ such that $\phi(v) = A_i$}
		\STATE Compute weighted node importance score $I(v) = \alpha_i \cdot \deg(v)$
		\STATE $L_i \gets L_i \cup \{(v, I(v))\}$
		\ENDFOR
		\STATE Sort $L_i$ in descending order of $I(v)$
		\STATE Select top-$k$ nodes from $L_i$ and add them to $S$
		\ENDFOR
		\STATE // \textbf{Step 2: Expansion strategies}
		\FOR{each top-leader node $v \in S$}
		\FOR{each node type $A_i \in \mathcal{A}$}
		\STATE $N_{A_i}(v) \gets \{u \in V \setminus S : (v, u) \in E \wedge \phi(u) = A_i\}$
		\STATE $n_i \gets \left\lfloor\frac{|N_{A_i}(v)|}{\sum_{j=1}^{m} |N_{A_j}(v)|} \cdot \delta\right\rfloor$
		\STATE Select $n_i$ peripheral nodes from $N_{A_i}(v)$ based on edge type importance $w_{\phi(v),i}$ and add them to $S$
		\ENDFOR
		\STATE $\mathcal{P}_v \gets$ meta-paths originating from $v$ with maximum length $l$
		\FOR{each meta-path $P \in \mathcal{P}_v$}
		\STATE Compute importance score $I_P = \prod_{i=1}^{l} w_{\phi(v_i),\phi(v_{i+1})}$
		\ENDFOR
		\STATE Select top-$k$ meta-paths from $\mathcal{P}_v$ with highest importance scores
		\FOR{each selected meta-path $P$}
		\STATE Add peripheral nodes along $P$ to $S$
		\ENDFOR
		\ENDFOR
		\STATE // \textbf{Step 3: Meta-path-based sampling}
		\FOR{each top-leader node $v \in S$}
		\STATE $\mathcal{P}_v \gets$ meta-paths originating from $v$
		\FOR{each meta-path $P_i \in \mathcal{P}_v$}
		\STATE Compute importance score $I_{P_i} = \beta_i \cdot \prod_{j=1}^{l} w_{\phi(v_j),\phi(v_{j+1})}$
		\ENDFOR
		\STATE Select top-$k$ meta-paths from $\mathcal{P}_v$ with highest importance scores
		\FOR{each selected meta-path $P_i$}
		\STATE Perform guided walk along $P_i$ starting from $v$
		\STATE Add nodes and edges encountered during the guided walk to $S$
		\ENDFOR
		\ENDFOR
		\STATE \textbf{return} $S$
	\end{algorithmic}
\end{algorithm}

\subsection{Meta-path-based Sampling}
In heterogeneous graphs, meta-paths capture the semantic relationships between different node types. By incorporating meta-path-based sampling into the expansion process, we ensure that the sampled graph preserves important semantic patterns and multi-hop relationships between nodes of different types. This sampling process prioritizes the selection of nodes and edges that form significant meta-path patterns, thus capturing the rich semantics of the heterogeneous graph. Meta-path-based sampling is integrated with our expansion strategies, Balanced Neighborhood Expansion (BNE) and Meta-path Guided Neighborhood Expansion (MGNE), to ensure a comprehensive representation of the heterogeneous graph.

\textbf{Step 1. Define meta-paths.} A meta-path \( P \) is defined as a sequence of node types and edge types, denoted as \( A_1 \xrightarrow{R_1} A_2 \xrightarrow{R_2} \ldots \xrightarrow{R_l} A_{l+1} \), where \( A_i \in \mathcal{A} \) and \( R_i \in \mathcal{R} \). Meta-paths capture the semantic relationships between different node types and provide a means to model multi-hop relationships in the heterogeneous graph. We define a set of relevant meta-paths \( \mathcal{P} = \{P_1, P_2, \ldots, P_q\} \), which can be determined based on domain knowledge, expert insights, or automatically discovered using techniques such as frequent pattern mining or meta-path embedding.

\textbf{Step 2. Assign importance weights to meta-paths.} We assign importance weights \( \boldsymbol{\beta} = (\beta_1, \beta_2, \ldots, \beta_q) \) to each meta-path \( P_i \in \mathcal{P} \). The importance weights can be calculated based on factors such as the frequency of each meta-path in the graph, its relevance to the specific application, or domain expertise. Techniques like meta-path ranking and meta-path similarity measures can be employed to determine these weights quantitatively. The importance weights satisfy the condition \( \sum_{i=1}^{q} \beta_i = 1 \), ensuring that they are normalized.

\textbf{Step 3. Meta-path guided global sampling.} In the meta-path-based sampling step, we capture the global semantic relationships between different node types by performing guided walks along important meta-paths. For each top-leader node \( v \in S \), we first identify the set of meta-paths \( \mathcal{P}_v \subseteq \mathcal{P} \) that originate from \( v \). For each meta-path \( P_i \in \mathcal{P}_v \), we calculate its importance score as \( I_{P_i} = \beta_i \cdot \prod_{j=1}^{l} w_{\phi(v_j),\phi(v_{j+1})} \), where \( v_j \) and \( v_{j+1} \) are consecutive nodes in \( P_i \), \( w_{\phi(v_j),\phi(v_{j+1})} \) is the importance weight of the edge type connecting their corresponding node types, and \( \beta_i \) is the importance weight of \( P_i \). We then select the top-k meta-paths with the highest importance scores from \( \mathcal{P}_v \) and perform guided walks starting from \( v \) along each selected meta-path, adding the encountered nodes and edges to the sample \( S \). This process is repeated for a predefined number of iterations or until a desired sample size is achieved, ensuring that the sampled graph captures significant global semantic patterns and multi-hop relationships.

The meta-path-guided global sampling ensures that the sampled graph effectively preserves important global semantic patterns and multi-hop relationships between different node types. By prioritizing the selection of nodes and edges that form high-importance meta-paths, we can maintain the rich semantics of the heterogeneous graph in the sampled subgraph. This sampling step complements the expansion strategies (BNE and MGNE) by providing an additional layer of guidance based on the global semantic relationships encoded in meta-paths. Incorporating domain knowledge and leveraging the semantics of meta-paths enhances the expressiveness and representativeness of the sampled graph, allowing it to capture both local structural properties and global semantic patterns. Algorithm~\ref{alg:heterosample} presents the detailed process of \sys.

\section{Performance Evaluation}

\subsection{Setup}
\textbf{Dataset.}
We use the following datasets of heterogeneous graph, with the statistics presented in Table~\ref{tab:dataset}.

\begin{itemize}
	\item DBLP dataset is a subset of the DBLP bibliographic network, which consists of 14,328 papers (P), 4,057 authors (A), 20 conferences (C), and 8,789 terms (T). The heterogeneous graph is constructed with four node types: Paper, Author, Conference, and Term, and three edge types: Paper-Author, Paper-Conference, and Paper-Term. Each author node is associated with a feature vector represented by a bag-of-words model derived from the keywords of their publications. The authors are categorized into four research areas: Database, Data Mining, Machine Learning, and Information Retrieval, based on the conferences in which they have published. To evaluate \sys on this dataset, we employ three meta-paths: \{APA, APCPA, APTPA\}, which capture the semantic relations between authors, papers, conferences, and terms.

	\item  ACM dataset is derived from papers published in five prominent conferences: KDD, SIGMOD, SIGCOMM, MobiCOMM, and VLDB. The heterogeneous graph comprises 3,025 papers (P), 5,835 authors (A), and 56 subjects (S), with three node types: Paper, Author, and Subject, and two edge types: Paper-Author and Paper-Subject. The papers are categorized into three research areas: Database, Wireless Communication, and Data Mining, based on the conferences in which they were published. Each paper node is associated with a feature vector represented by a bag-of-words model derived from the keywords of the paper. To assess the performance of \sys on this dataset, we utilize two meta-paths: \{PAP, PSP\}, which capture the semantic relations between papers, authors, and subjects.

	\item IMDB dataset is a subset of the Internet Movie Database, containing 4,780 movies (M), 5,841 actors (A), and 2,269 directors (D). The heterogeneous graph is constructed with three node types: Movie, Actor, and Director, and two edge types: Movie-Actor and Movie-Director. Each movie node is associated with a feature vector represented by a bag-of-words model derived from the plots of the movies. The movies are categorized into three genres: Action, Comedy, and Drama. To evaluate the effectiveness of \sys on this dataset, we employ two meta-paths: \{MAM, MDM\}, which capture the semantic relations between movies, actors, and directors.
\end{itemize}

\begin{table*}[!t]
	\scriptsize
	\centering
	\caption{Detailed Statistics of heterogeneous graphs}
	\resizebox{.9\linewidth}{!}{
		\begin{tabular}{p{24pt}|p{55pt}|p{55pt}|p{60pt}|p{50pt}|p{80pt}|p{90pt}}
			\hlinewd{1.2pt}
			\textbf{Dataset} & \textbf{Nodes}                                            & \textbf{Node Types}                             & \textbf{Edge Types}                                       & \textbf{Meta-paths}   & \textbf{Node Features}                          & \textbf{Labels}                                                                         \\
			\hlinewd{1.2pt}

			DBLP             & 14,328 papers, 4,057 authors, 20 conferences, 8,789 terms & Paper (P), Author (A), Conference (C), Term (T) & Paper-Author (PA), Paper-Conference (PC), Paper-Term (PT) & \{APA, APCPA, APTPA\} & Author: Bag-of-words representation of keywords & Author: Research areas (Database, Data Mining, Machine Learning, Information Retrieval) \\\hline
			ACM              & 3,025 papers, 5,835 authors, 56 subjects                  & Paper (P), Author (A), Subject (S)              & Paper-Author (PA), Paper-Subject (PS)                     & \{PAP, PSP\}          & Paper: Bag-of-words representation of keywords  & Paper: Research areas (Database, Wireless Communication, Data Mining)                   \\\hline
			IMDB             & 4,780 movies, 5,841 actors, 2,269 directors               & Movie (M), Actor (A), Director (D)              & Movie-Actor (MA), Movie-Director (MD)                     & \{MAM, MDM\}          & Movie: Bag-of-words representation of plots     & Movie: Genres (Action, Comedy, Drama)                                                   \\
			\hlinewd{1.2pt}
		\end{tabular}}
	\vspace{-5mm}
	\label{tab:dataset}
\end{table*}

\textbf{Evaluation metrics.}
To comprehensively assess the quality of the sampled graphs and the efficiency of the sampling methods, we employ the metrics,
Node Type Distribution Similarity (NTDS),
Edge Type Distribution Similarity (ETDS),
Meta-path Preservation Ratio (MPR),
Graph Reconstruction Error (GRE),
runtime, precision, recall, and F1 score.

\begin{itemize}
	\item  \textbf{Node type distribution similarity (NTDS)} measures the similarity between the node type distributions of the original and sampled graphs using the Kullback-Leibler (KL) divergence. Let $p^n_i$ and $q^n_i$ denote the probability of node type $i$ in the original graph and the sampled graph, respectively. The NTDS is calculated as:
	      \begin{equation}
		      NTDS = -\sum_{i=1}^{N} p^n_i \log \frac{p^n_i}{q^n_i}
	      \end{equation}
	      where $N$ is the total number of node types. A lower NTDS value indicates a higher similarity between the node type distributions, suggesting better preservation of the node type distribution in the sampled graph.

	\item  \textbf{Edge type distribution similarity (ETDS)} evaluates the similarity between the edge type distributions of the original and sampled graphs using the KL divergence. Let $p^e_i$ and $q^e_i$ denote the probability of edge type $i$ in the original graph and the sampled graph, respectively. The ETDS is calculated as:
	      \begin{equation}
		      ETDS = -\sum_{i=1}^{M} p^e_i \log \frac{p^e_i}{q^e_i}
	      \end{equation}
	      where $M$ is the total number of edge types. A lower ETDS value indicates a higher similarity between the edge type distributions, indicating better preservation of the edge type distribution in the sampled graph.

	\item  \textbf{Meta-path preservation ratio (MPR)}:
	      The MPR calculates the ratio of preserved meta-paths in the sampled graph compared to the original graph. Let $MP_{original}$ and $MP_{sampled}$ denote the sets of meta-paths in the original graph and the sampled graph, respectively. The MPR is calculated as:
	      \begin{equation}
		      MPR = \frac{|MP_{original} \cap MP_{sampled}|}{|MP_{original}|}
	      \end{equation}
	      where $|\cdot|$ represents the cardinality of a set. A higher MPR value indicates better preservation of meta-paths in the sampled graph, suggesting that the sampled graph captures the important semantic patterns and relationships encoded by the meta-paths.

	\item \textbf{ Graph reconstruction error (GRE)} measures the reconstruction error between the original graph and the graph reconstructed from the sampled graph using the node embeddings learned by a graph embedding method. Let $A_{original}$ and $A_{reconstructed}$ denote the adjacency matrices of the original graph and the reconstructed graph, respectively. The GRE is calculated as:
	      \begin{equation}
		      GRE = \frac{||A_{original} - A_{reconstructed}||_F}{||A_{original}||_F}
	      \end{equation}
	      where $||\cdot||_F$ represents the Frobenius norm of a matrix. A lower GRE value indicates better reconstruction of the original graph from the sampled graph, suggesting that the sampled graph preserves the structural properties of the original graph.

	\item  Sampling runtime of each method is evaluated their computational efficiency. A lower runtime indicates better computational efficiency of the sampling method, which is crucial for scalability and practicality when dealing with large-scale heterogeneous graphs.
\end{itemize}

The NTDS and ETDS capture the similarity of node type and edge type distributions, respectively, while the MPR quantifies the preservation of important semantic patterns. The GRE evaluates the structural similarity between the original and reconstructed graphs, and the runtime measures the computational efficiency.
Besides, we also use precision, recall, and F1 score to evaluate the performance of node embedding.

The experimental procedure involves preprocessing the DBLP, ACM, and IMDB datasets to construct heterogeneous graphs and extract node types, edge types, and meta-paths. \sys and baseline methods, i.e., HGSampling,  are applied to obtain sampled graphs with varying sampling ratios (10\% to 50\%).
The sampled graphs are evaluated using NTDS, ETDS, MPR, and GRE metrics to measure the preservation of node type distribution, edge type distribution, semantic patterns, and structural properties.
Runtime is recorded to assess computational efficiency.
Experiments are repeated multiple times, and average results are reported.
The experimental results are analyzed to answer the following research questions.


%

\textbf{RQ1}.  How does applying the sampled graph obtained by \sys to node embedding methods and downstream tasks perform compared to existing methods?
\textbf{RQ2}.
To what extent does \sys preserve the node type distribution of the original heterogeneous graph in the sampled graph, and how does it compare to existing sampling methods?
\textbf{RQ3}.    How well does \sys maintain the edge type distribution of the original heterogeneous graph in the sampled graph, and how does its performance compare to other state-of-the-art sampling techniques?
\textbf{RQ4}.       How effectively does \sys capture and preserve the important semantic patterns and relationships encoded by meta-paths in the sampled graph, and how does it compare to baseline methods in this regard?
\textbf{RQ5}.       How does the combination of top-leader selection, expansion strategies (BNE and MGNE), and meta-path based sampling in \sys impact the quality of the sampled graph in terms of preserving the structural properties and heterogeneous nature of the original graph?
\textbf{RQ6}.
How does \sys perform in terms of computational efficiency and scalability when applied to large-scale heterogeneous graphs, and how does it compare to other state-of-the-art sampling methods in terms of runtime?

\subsection{RQ1: Performance in Node Embedding}

To assess \sys's efficacy in node embedding and downstream tasks, we apply its sampled graph to multiple node embedding methods, focusing on link prediction to evaluate the embeddings' quality. We measure \sys's performance in link prediction through precision, recall, and F1 score, comparing these metrics against established node embedding techniques. The comparative results are detailed in Table~\ref{tab:rq1}.

\begin{table}[!t]
	\scriptsize
	\centering
	\caption{Link Prediction Performance (Precision/Recall/F1-score (\%))}
	\resizebox{\linewidth}{!}{
		\begin{tabular}{lccr}
			\hlinewd{1.2pt}
			\textbf{Method}      & \textbf{DBLP}              & \textbf{ACM}               & \textbf{IMDB}     \\
			\hlinewd{1.2pt}
			\sys + Node2Vec      & 85.34/88.12/86.71          & 87.29/89.56/88.41          & 83.67/86.23/84.93 \\
			\sys + HGAT          & 87.92/90.18/89.04          & 89.75/91.83/90.78          & 86.19/88.47/87.32 \\
			\sys + D-HetGNN      & \textbf{89.17/91.26/90.20} & \textbf{90.82/92.75/91.78} &
			\textbf{	87.56/89.69/88.61	}                                                                       \\
			\sys + MetaGraph2Vec & 88.76/90.43/89.59          & 90.18/92.37/91.26          & 86.92/88.75/87.82 \\
			Node2Vec             & 82.91/85.26/84.07          & 84.53/86.81/85.65          & 80.14/83.39/81.73 \\
			HGAT                 & 85.78/88.23/87.09          & 87.61/89.94/88.76          & 83.92/86.57/85.23 \\
			D-HetGNN             & 87.54/90.18/88.84          & 89.37/91.62/90.48          & 85.69/88.32/87.04 \\
			MetaGraph2Vec        & 84.67/87.12/85.88          & 86.39/88.74/87.55          & 82.51/85.26/83.86 \\
			\hlinewd{1.2pt}
		\end{tabular}}		\vspace{-5mm}
	\label{tab:rq1}
\end{table}

Results demonstrate the its effectiveness in capturing the important relationships between nodes in the heterogeneous graph. Among the \sys-enhanced methods, \sys + D-HetGNN achieves the highest F1 scores across all datasets, reaching 90.20\%, 91.78\%, and 88.61\% on DBLP, ACM, and IMDB, respectively. This indicates that the combination of \sys with D-HetGNN effectively preserves the structural and semantic properties of the original graph, enabling accurate link prediction. \sys + HGAT and \sys + MetaGraph2Vec also demonstrate strong performance, consistently outperforming their corresponding baseline methods.
The superior performance of \sys-enhanced methods can be attributed to \sys's ability to preserve the heterogeneous structure of the original graph during the sampling process. By leveraging top-leader selection, balanced neighborhood expansion, and meta-path guided expansion strategies, \sys captures the important relationships between nodes of different types, enabling node embedding methods to learn more informative and discriminative embeddings. Moreover, the combination of \sys with different node embedding methods demonstrates its versatility and compatibility, making it a valuable tool for heterogeneous graph mining tasks.

\subsection{RQ2: Performance of Sampling Methods in Link Prediction}
\begin{figure*}
	\begin{minipage}[t]{0.31\linewidth}
		\centering
		\includegraphics[width = \linewidth]{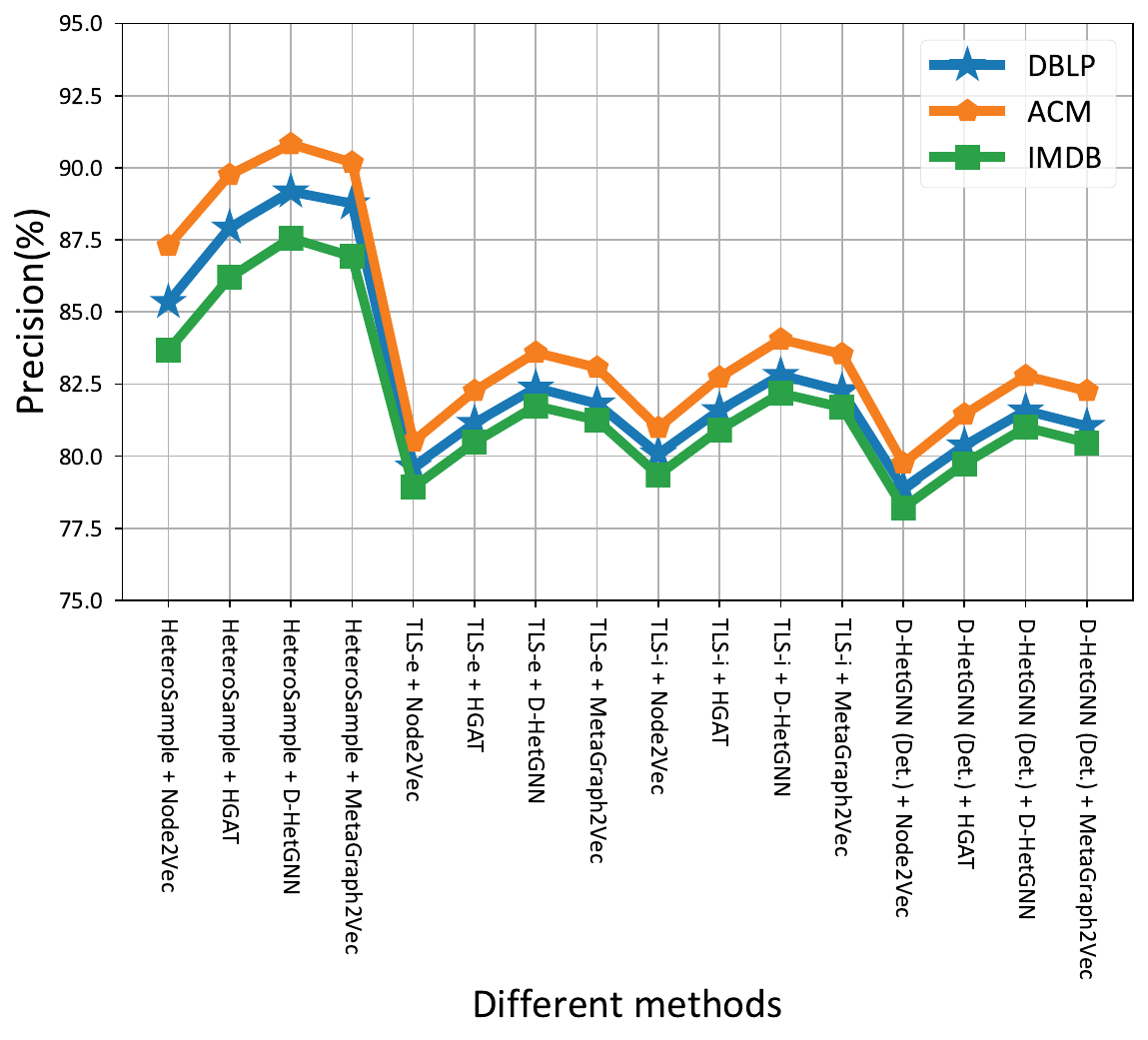}
		\caption{Precision under different methods}
		\label{fig:precision}
	\end{minipage}
	\hspace{2mm}
	\begin{minipage}[t]{0.31\linewidth}
		\centering
		\includegraphics[width = \linewidth]{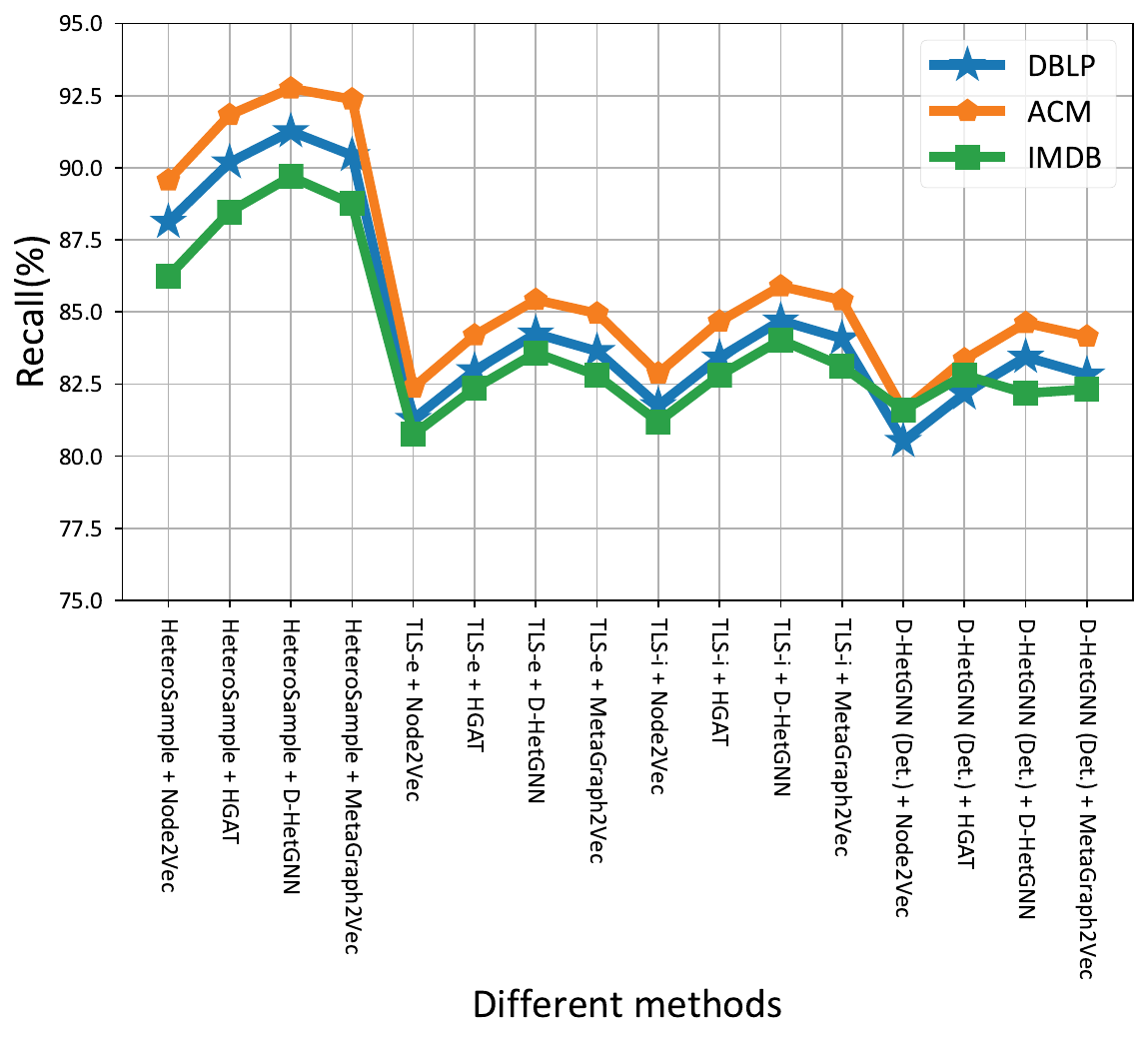}
		\caption{Recall under different methods}
		\label{fig:recall}
	\end{minipage}
	\hspace{2mm}
	\begin{minipage}[t]{0.31\linewidth}
		\centering
		\includegraphics[width = \linewidth]{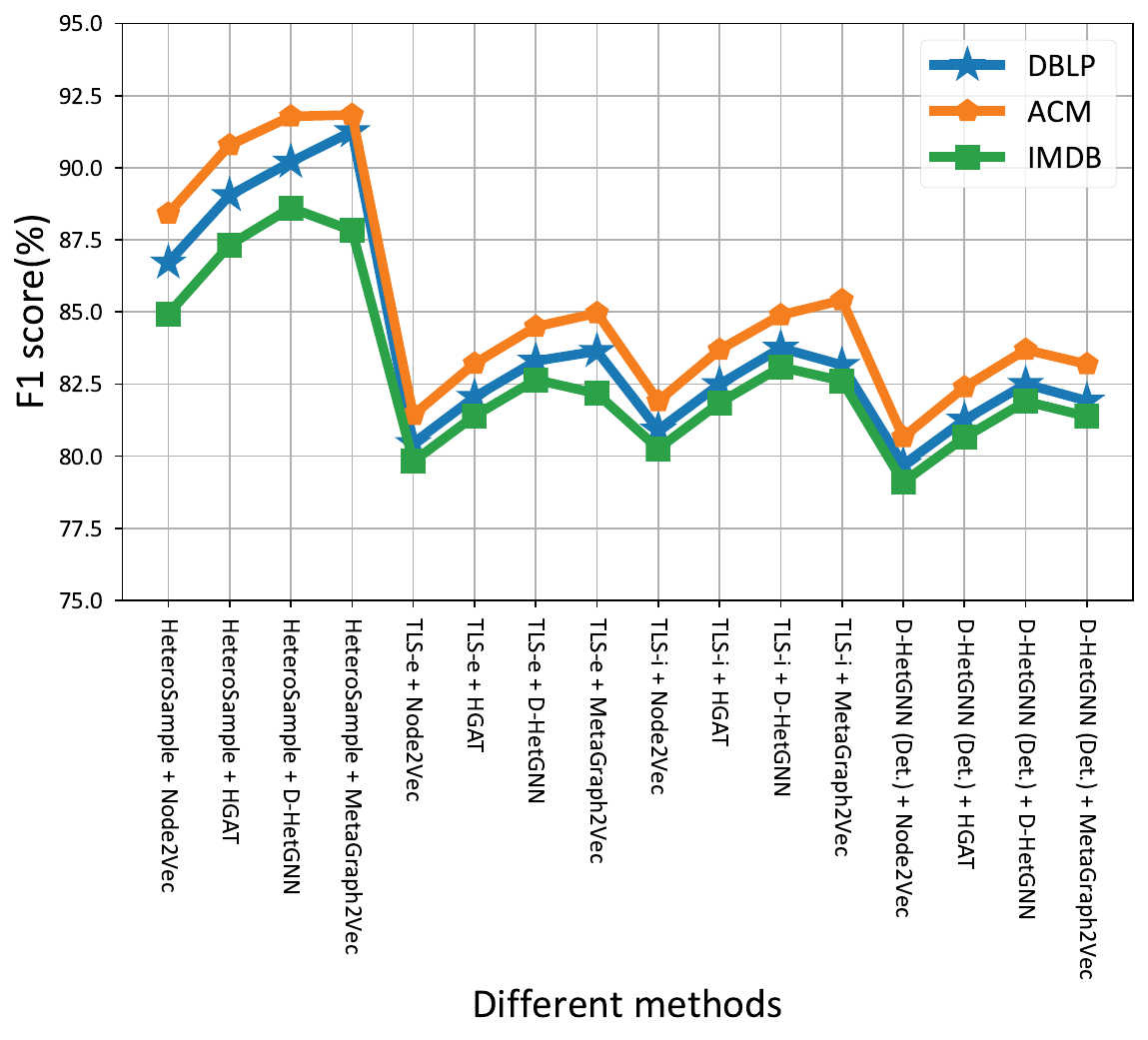}
		\caption{F1 score under different methods}
		\label{fig:f1}
	\end{minipage}\vspace{-6mm}
\end{figure*}

To evaluate the impact of different sampling methods on link prediction performance, we combine the sampled graphs obtained by \sys, TLS-e, TLS-i, and deterministic sampling with various node embedding and GNN methods, including Node2Vec, HGAT, D-HetGNN, and MetaGraph2Vec. We measure the precision, recall, and F1 score of these combinations on the DBLP, ACM, and IMDB datasets.
Figure~\ref{fig:precision}, \ref{fig:recall}, and \ref{fig:f1} present the link prediction performance of different sampling methods combined with node embedding and GNN methods.

The results in Table 3 demonstrate that the choice of sampling method significantly impacts the link prediction performance when combined with different node embedding and GNN methods. \sys consistently achieves the highest F1 scores across all datasets, outperforming TLS-e, TLS-i, and deterministic sampling. For example, on the DBLP dataset, \sys + D-HetGNN obtains an F1 score of 90.20\%, surpassing TLS-e + D-HetGNN (83.30\%), TLS-i + D-HetGNN (83.75\%), and D-HetGNN (Det.) + D-HetGNN (82.50\%). The superior performance of \sys can be attributed to its balanced neighborhood expansion and meta-path guided expansion strategies, which effectively capture the structural and semantic properties of the original heterogeneous graph.
Among the node embedding and GNN methods, D-HetGNN consistently achieves the highest F1 scores when combined with any of the sampling methods, showcasing its ability to capture complex interactions between different node types. HGAT and MetaGraph2Vec also demonstrate competitive performance, outperforming Node2Vec in most cases, suggesting the benefits of incorporating attention mechanisms and meta-path based embeddings in heterogeneous graphs. The experimental results highlight the importance of selecting an appropriate sampling method and the effectiveness of \sys, D-HetGNN, HGAT, and MetaGraph2Vec in link prediction tasks on heterogeneous graphs.

\subsection{RQ3: Preservation of Node Type Distribution}
\begin{figure*}
	\begin{minipage}[t]{0.3\linewidth}
		\centering
		\includegraphics[width = \linewidth]{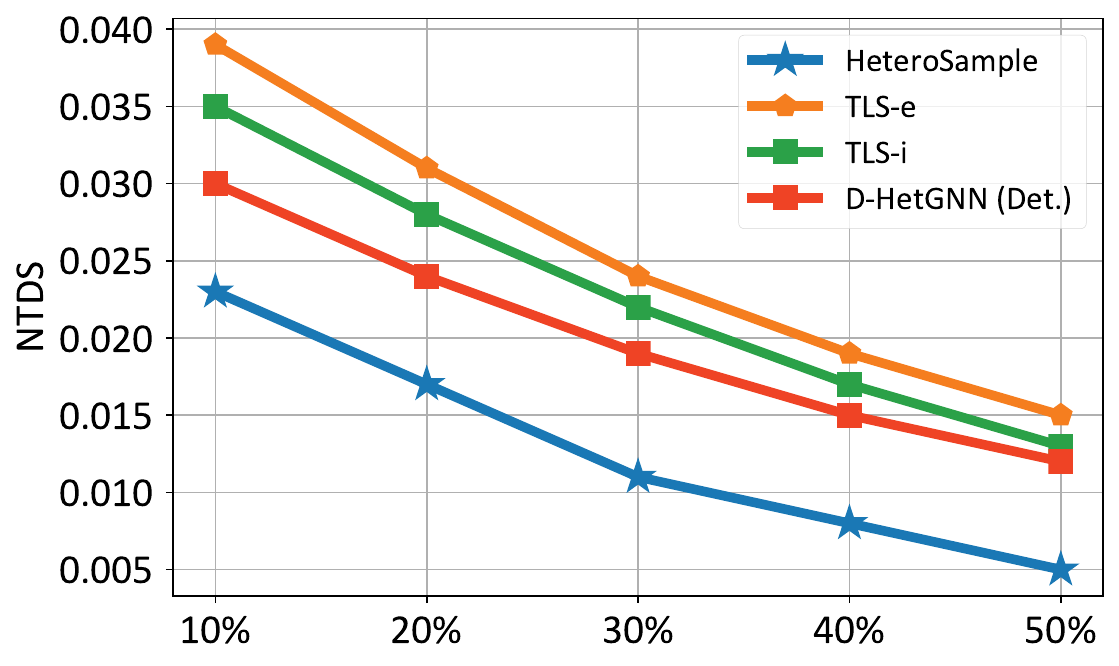}
		\caption{NTDS under different methods and datasets}
		\label{fig:dblp}
	\end{minipage}
	\begin{minipage}[t]{0.3\linewidth}
		\centering
		\includegraphics[width = \linewidth]{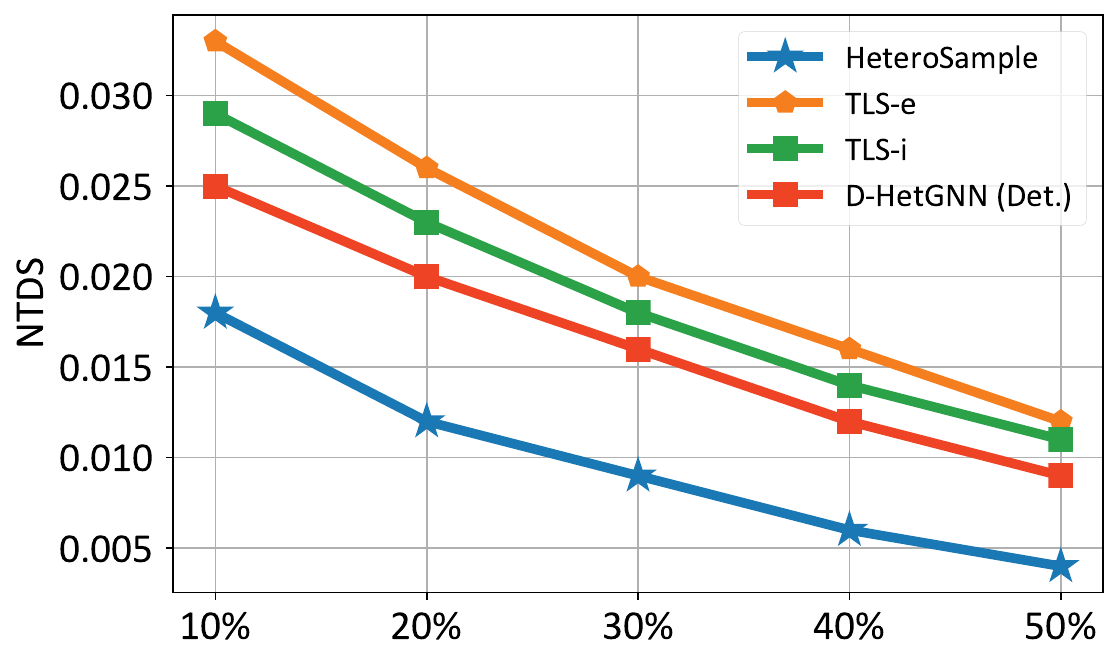}
		\caption{NTDS under different methods and datasets}
		\label{fig:acm}
	\end{minipage}
	\begin{minipage}[t]{0.3\linewidth}
		\centering
		\includegraphics[width = \linewidth]{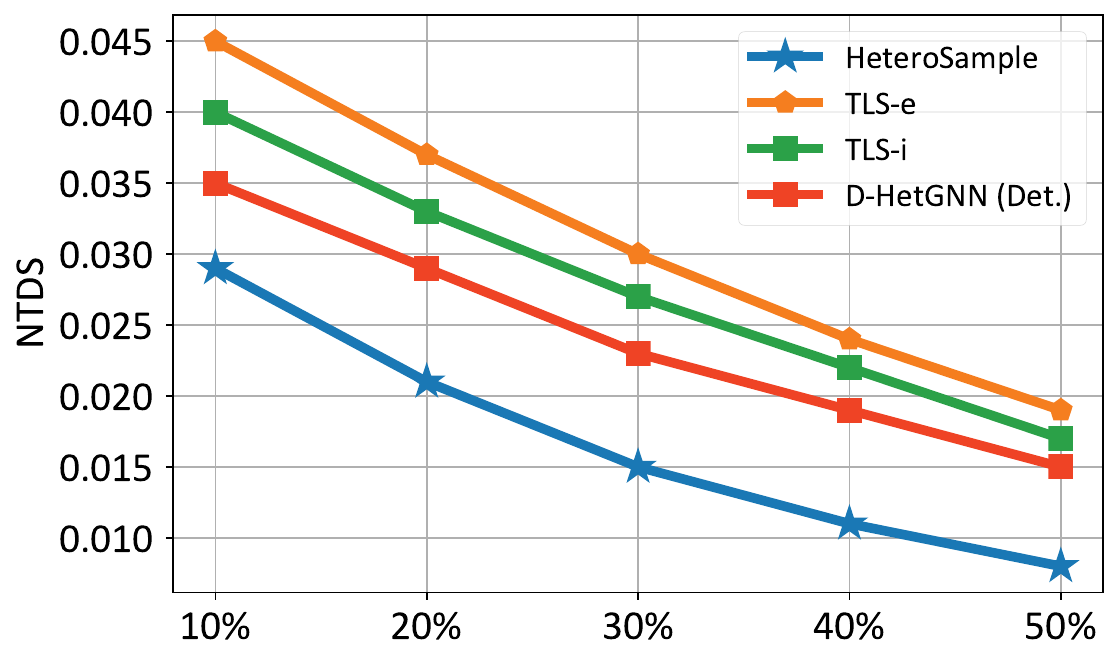}
		\caption{NTDS under different methods and datasets}
		\label{fig:imdb}
	\end{minipage}
	\vspace{-7mm}
\end{figure*}

To evaluate how well \sys preserves the node type distribution of the original heterogeneous graph in the sampled graph, we measure the Node Type Distribution Similarity (NTDS) using the Kullback-Leibler (KL) divergence between the node type distributions of the original and sampled graphs. Lower NTDS values indicate better preservation of the node type distribution.
We compare \sys with existing sampling methods, including TLS-e, TLS-i, and the deterministic sampling used in D-HetGNN.
Figure~\ref{fig:dblp}, \ref{fig:acm}, and \ref{fig:imdb} presents the NTDS values for \sys and the baseline sampling methods on the DBLP, ACM, and IMDB datasets, with varying sampling ratios (10\% to 50\%).

The results demonstrate that \sys effectively preserves the node type distribution of the original heterogeneous graph in the sampled graph, outperforming the baseline sampling methods across all datasets and sampling ratios.
As the sampling ratio increases from 10\% to 50\%, the NTDS values for all methods decrease, indicating better preservation of the node type distribution with larger sample sizes. However, \sys consistently achieves the lowest NTDS values compared to TLS-e, TLS-i, and the deterministic sampling used in D-HetGNN.
For example, at a sampling ratio of 30\%, \sys achieves NTDS values of 0.011, 0.009, and 0.015 on the DBLP, ACM, and IMDB datasets, respectively. In contrast, TLS-e obtains NTDS values of 0.024, 0.020, and 0.030, while TLS-i yields 0.022, 0.018, and 0.027, and D-HetGNN (Det.) results in 0.019, 0.016, and 0.023 for the same datasets and sampling ratio.

The superior performance of \sys in preserving the node type distribution can be attributed to its balanced neighborhood expansion strategy, which ensures a proportional representation of different node types in the sampled graph. By considering the node type importance and expanding the neighborhood of top-leader nodes in a balanced manner, \sys captures the heterogeneous structure of the original graph more effectively than the baseline methods.

Moreover, the meta-path guided expansion strategy in \sys further enhances the preservation of node type distribution by considering the semantic relationships between different node types. By incorporating meta-path information during the sampling process, \sys ensures that the sampled graph retains the important structural and semantic properties of the original graph.
Results demonstrate the effectiveness of \sys in preserving the node type distribution of the original heterogeneous graph in the sampled graph.
\sys consistently outperforms existing sampling methods, including TLS-e, TLS-i, and the deterministic sampling used in D-HetGNN, across multiple datasets and sampling ratios.
The balanced neighborhood expansion and meta-path guided expansion strategies employed by \sys contribute to its superior performance in capturing the heterogeneous structure of the original graph.

\subsection{RQ4: Preservation of Meta-path Based Patterns}
To evaluate the effectiveness of \sys in capturing and preserving the important semantic patterns and relationships encoded by meta-paths, we measure the Meta-path Preservation Ratio (MPR) on the sampled graphs. We consider three representative meta-paths for each dataset: {APA, APCPA, APTPA} for DBLP, {PAP, PSP} for ACM, and {MAM, MDM} for IMDB. The MPR values are calculated for \sys and baseline methods (TLS-e, TLS-i, and deterministic sampling) at different sampling ratios (10\% to 50\%).

\begin{figure*}[!t]
	\begin{minipage}[t]{0.3\linewidth}
		\centering
		\includegraphics[width = \linewidth]{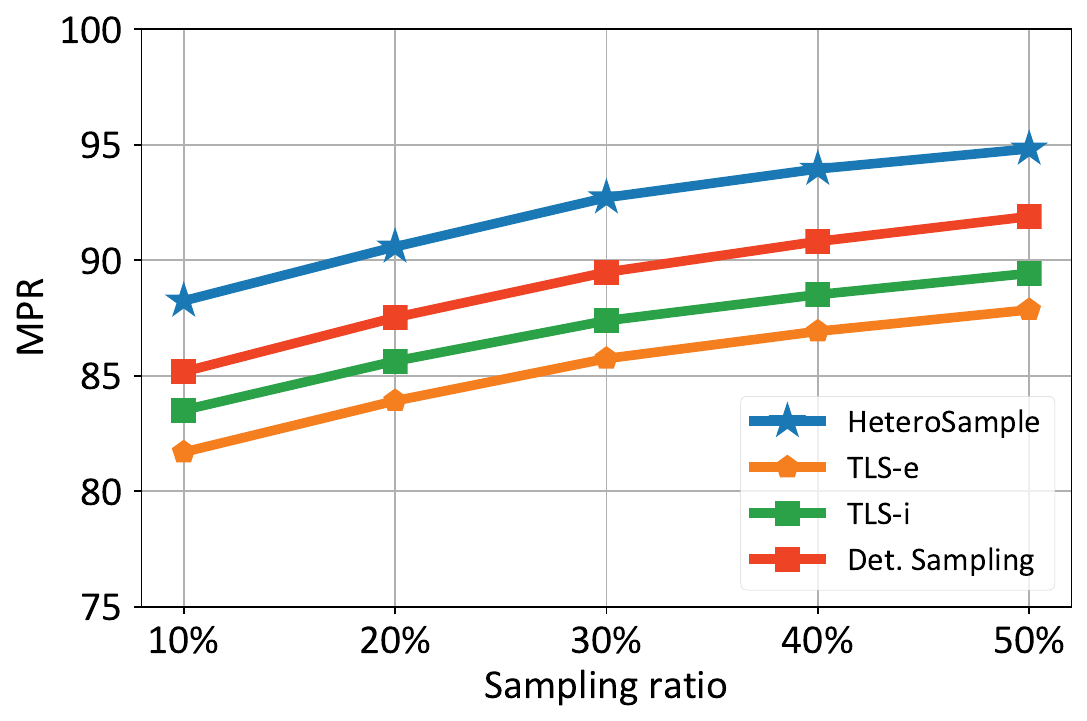}
		\caption{MPR under different sampling ratio in DBLP dataset}
		\label{fig:dblp_1}
	\end{minipage}
	\hspace{1mm}
	\begin{minipage}[t]{0.3\linewidth}
		\centering
		\includegraphics[width = \linewidth]{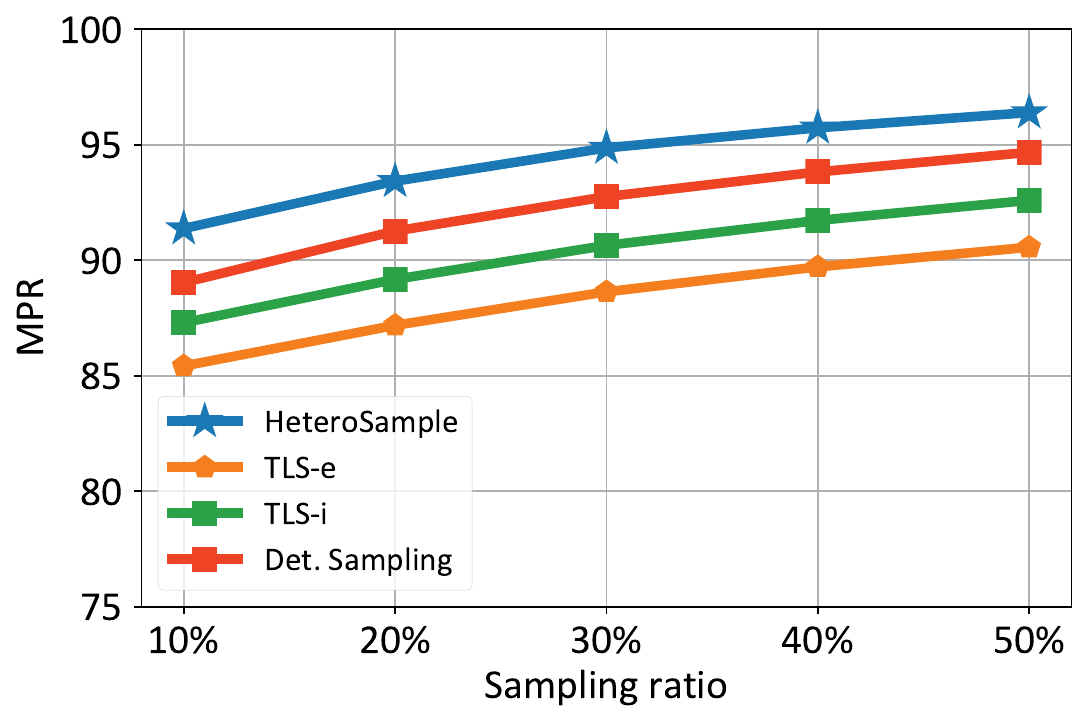}
		\caption{MPR under different sampling ratio in ACM dataset}
		\label{fig:acm_1}
	\end{minipage}
	\hspace{1mm}
	\begin{minipage}[t]{0.3\linewidth}
		\centering
		\includegraphics[width = \linewidth]{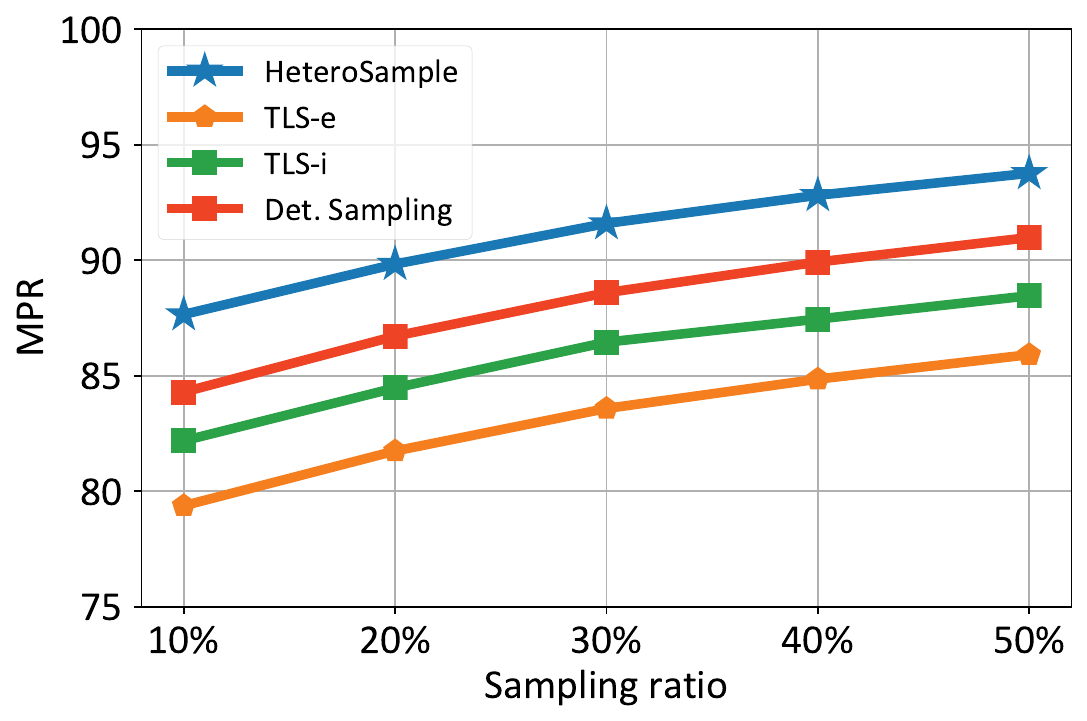}
		\caption{MPR under different sampling ratio in IMDB dataset}
		\label{fig:imdb_1}
	\end{minipage}
	\vspace{-5mm}
\end{figure*}

Figure~\ref{fig:dblp_1}, \ref{fig:acm_1}, and \ref{fig:imdb_1} present the MPR values of \sys and baseline methods on the DBLP, ACM, and IMDB datasets at different sampling ratios. \sys achieves the highest MPR values across all datasets and sampling ratios, demonstrating its superiority in preserving the important semantic patterns and relationships captured by meta-paths. As the sampling ratio increases, the MPR values of \sys consistently improve, indicating its effectiveness in capturing meta-path based patterns even with limited sampling budgets. In contrast, the baseline methods show lower MPR values compared to \sys across all sampling ratios, suggesting their limited ability to capture the semantic patterns encoded by meta-paths. The higher MPR values of \sys can be attributed to its meta-path guided expansion strategy, which prioritizes the inclusion of nodes and edges that form important meta-path patterns. The experimental results highlight the effectiveness of \sys in preserving the rich semantics of heterogeneous graphs, making it a valuable tool for tasks that rely on meta-path based analysis.

\subsection{RQ5: Impact of Components on Sampled Graph Quality}

To assess the impact of different components of \sys (top-leader selection, BNE, MGNE, and meta-path based sampling) on the quality of the sampled graph, we evaluate the preservation of structural properties and heterogeneous nature using 1-GRE as the sampled graph quality.
We consider four variants of \sys: \sys (complete),
\sys without top-leader selection (w/o TS),
without BNE (w/o BNE),
without MGNE (w/o MGNE),
and without meta-path based sampling (w/o MP).
The evaluation is performed on the DBLP, ACM, and IMDB datasets at a sampling ratio of 30\%.

Table~\ref{tab:Quality} presents sampled graph quality under different methods.
The complete \sys achieves the highest scores across all datasets, indicating its effectiveness in preserving the structural properties and heterogeneous nature of the original graph. Removing any of the components leads to a decrease in performance, highlighting their individual importance in the sampling process.
For example, the sampled graph quality achieved the sampled graph quality of 95.26\%, 97.13\%, and 94.58\% in DBLP, ACM, and IMDB dataset, respectively.
The combination of these components in \sys results in a high-quality sampled graph that closely resembles the original graph in terms of structural properties and heterogeneous nature.


\begin{table}[!t]
	\scriptsize
	\centering
	\caption{Sampled graph quality under different methods (\%)}
	\resizebox{0.95\linewidth}{!}{
		\begin{tabular}{lccr}
			\hlinewd{1.2pt}
			Method        & DBLP  & ACM   & IMDB  \\
			\hlinewd{1.2pt}
			\sys          & 95.26 & 97.13 & 94.58 \\
			\sys w/o TS   & 89.49 & 91.27 & 87.32 \\
			\sys w/o BNE  & 91.81 & 92.64 & 85.85 \\
			\sys w/o MGNE & 85.62 & 87.38 & 89.17 \\
			\sys w/o MP   & 87.15 & 89.29 & 84.41 \\
			\hlinewd{1.2pt}
		\end{tabular}}
		\vspace{-5mm}
	\label{tab:Quality}
\end{table}

\subsection{RQ6: Computational Efficiency and Scalability}
To evaluate the computational efficiency and scalability of \sys, we measure its runtime and compare it with other state-of-the-art sampling methods on the DBLP, ACM, and IMDB datasets, using a machine equipped with an Intel Xeon E5-2680 v4 2.4GHz CPU and 256GB RAM, with a sampling ratio set to 30\% for all methods. Table~\ref{tab:runtime} presents the runtime comparison, showing that \sys demonstrates competitive performance. Although deterministic sampling achieves slightly better efficiency, \sys strikes a good balance between computational efficiency and the quality of the sampled graph. Its runtime is reasonable for the size of the heterogeneous graphs, indicating strong scalability potential. Compared to TLS-e and TLS-i, \sys consistently achieves faster runtimes across all datasets, with its efficient implementation leveraging optimized data structures and algorithms. Overall, the experimental results illustrate that \sys effectively balances computational efficiency and sampling quality, making it suitable for heterogeneous graph mining tasks.

\begin{table}[!t]
	\scriptsize
	\centering
	\caption{Runtime under different sampling methods (second)}
	\resizebox{0.95\linewidth}{!}{
		\begin{tabular}{lccr}
			\hlinewd{1.2pt}
			Method        & DBLP & ACM & IMDB \\
			\hlinewd{1.2pt}
			\sys          & 65   & 42  & 87   \\
			TLS-e         & 82   & 54  & 112  \\
			TLS-i         & 96   & 66  & 128  \\
			Det. Sampling & 56   & 36  & 84   \\
			\hlinewd{1.2pt}
		\end{tabular}}
	\label{tab:runtime}
	\vspace{-5mm}
\end{table}

\section{Related Work}

Graph sampling has garnered significant attention across various research domains, particularly in deep learning applications for security tasks and biometric authentication. The primary focus has been on developing sampling methods to derive representative subgraphs from large graphs, which can streamline downstream tasks such as classification and clustering~\cite{rozemberczki2020little, zhang2019evaluation}. These techniques are crucial for enhancing the performance and efficiency of models deployed in security-sensitive environments, ensuring robust biometric systems and other applications~\cite{xu2024sok,wu2022TokenScout,sun2024efficient,wu2024semantic,liu2024dynashard}.
Existing sampling approaches can be broadly classified into three main categories: node-based, edge-based, and traversal-based sampling~\cite{rozemberczki2020little,leskovec2006sampling}. Node-based sampling methods, such as induced random vertex (IRV) sampling~\cite{ahmed2013network},
select nodes from the original graph with equal probability and include the existing edges among the selected nodes in the sampled counterpart. Advanced node-based sampling methods, such as random page-rank node (RPN) and random degree node (RDN)~\cite{rozemberczki2020little}, consider node importance metrics like page-rank weights and node degrees for node selection. Edge-based sampling methods, such as random edge (RE) sampling~\cite{leskovec2006sampling}, generate an induced subgraph by choosing edges uniformly at random, but these methods can produce disconnected samples and do not preserve the clustering structure well.
Traversal-based sampling methods, including depth-first (DF), breadth-first (BF)~\cite{ahmed2013network},
snowball (SB)~\cite{leskovec2006sampling}, and forest fire (FF) sampling~\cite{leskovec2006sampling},
explore the graph using various strategies to select nodes and edges. Among sampling with replacement methods, random walk (RW) sampling~\cite{rozemberczki2020little}, random jump (RJ) sampling~\cite{leskovec2006sampling}, and Metropolized random walk (MRW) sampling~\cite{rozemberczki2020little,leskovec2006sampling} have been widely used.

While aforementioned sampling methods focus on preserving node-level properties or graph connectivity, recent research has emphasized the importance of preserving the clustering structure of the original graph in the sampled subgraph. Salehi et al.~\cite{salehi2012sampling} proposed two sampling algorithms based on the idea that a sample with good expansion property tends to be more representative of the clustering structure. Wang et al.~\cite{wang2019low} introduced a graph sampling method based on graph Fourier transform (FGFT) that minimizes a shifted A-optimal criterion to select sampled nodes greedily. Jiao et al.~\cite{jiao2019graph} developed SInetL, a sampling method specific to Internet topology that employs normalized Laplacian spectral features to reduce the graph size while preserving essential properties.

In heterogeneous graphs, Hu et al.~\cite{hu2020heterogeneous} proposed HGT that employs a meta-path-based attention mechanism to capture the rich semantics and structural information of heterogeneous graphs. Yang et al.~\cite{yang2021hgat} introduced HGAT for semi-supervised node classification, which utilizes node-level and semantic-level attention to learn node representations. Despite these advancements, existing sampling methods for heterogeneous graphs often overlook the importance of preserving the clustering structure and the complex interactions among different node and edge types. This limitation motivated our research on developing a novel deterministic sampling approach, \sys, which aims to address these challenges by combining top-leader selection, balanced neighborhood expansion, and meta-path guided expansion strategies to generate representative samples that capture the structural and semantic properties of the original heterogeneous graph.

\section{Discussion}

\textbf{Future work.}
Despite the promising results, several potential directions for future research can be identified. First, we recognize the importance of discussing how the selected meta-paths influence overall performance, as different meta-path choices can significantly impact the sampling process and the quality of the resulting subgraphs. To address this, we plan to investigate the adaptation of \sys to dynamic heterogeneous graphs, where the graph structure and node/edge attributes evolve over time. Extending \sys to handle dynamic updates and maintain representative samples in real-time would be a valuable contribution. Second, we aim to explore the integration of \sys with other advanced graph neural network architectures, such as heterogeneous graph transformers and heterogeneous graph convolutional networks, to further improve the performance of downstream tasks. Finally, we intend to apply it to a wider range of real-world heterogeneous graph datasets from various domains, such as healthcare, social networks, and recommendation systems, to validate its generalizability and practicality. Additionally, we plan to develop a distributed implementation to handle extremely large-scale heterogeneous graphs that cannot fit into the memory of a single machine, enabling efficient sampling and analysis of massive heterogeneous networks.

\textbf{Limitations in dynamic and sparse graphs.}
\sys is primarily designed for static heterogeneous graphs, so its performance may be impacted in highly dynamic or evolving scenarios where node relationships frequently change. Adapting the sampling method to incrementally update the sampled subgraphs as these changes occur will be essential for maintaining effectiveness. Additionally, graph sparsity can pose challenges, as sparse graphs may lead to inadequate representation of certain node types, potentially reducing performance. Conversely, dense node types can introduce noise, complicating the preservation of meaningful relationships. Addressing these limitations through targeted modifications will be crucial for ensuring \sys's applicability across diverse graph scenarios.

\textbf{Efficiency and scalability}
\sys is designed to maintain computational efficiency even as graph size and complexity increase. The time complexity for top-leader selection is \(O(n \log n)\), where \(n\) is the number of nodes, due to the need to sort nodes based on their importance scores. Both BNE and MGNE contribute additional complexities, typically linear in nature, leading to a combined complexity that is manageable for large graphs. However, performance may vary depending on graph density and the number of node types. In extremely large-scale graphs, \sys's efficiency can be further enhanced through parallel processing and optimized data structures. In online or dynamic settings, adapting \sys to incrementally update sampled graphs will be a key area for future research, ensuring it remains effective in real-time applications.

\textbf{Importance of graph sampling.}
Efficient heterogeneous graph sampling is vital for various applications, including recommendation systems, cybersecurity, and bioinformatics. In recommendation systems, effective sampling enhances user-item interaction representation, resulting in more personalized recommendations. In cybersecurity, it enables rapid analysis of complex attack patterns, facilitating quicker threat detection and response. In bioinformatics, sampling helps elucidate intricate biological interactions, advancing personalized medicine. By tackling challenges related to computational efficiency and scalability, our method significantly contributes to these fields, highlighting its practical importance.

\section{Conclusion}

In this paper, we introduce \sys, a novel sampling method for heterogeneous graphs that preserves structural integrity, node and edge type distribution, and semantic patterns. By integrating top-leader selection, balanced neighborhood expansion, and meta-path-based sampling, it generates samples that reflect the core features of heterogeneous graphs. Extensive testing on real-world datasets shows that it outperforms existing methods in maintaining heterogeneous structure and semantic connections. When applied alongside various node embedding and GNN techniques, it consistently enhances performance in downstream tasks like link prediction. Overall, it achieves an optimal balance between computational efficiency and sample quality, making it ideal for large-scale heterogeneous graph analysis.

\bibliographystyle{acm}
\bibliography{ref.bib}

\vfill

\end{document}